# UUV's Hierarchical DE-based Motion Planning in a Semi Dynamic Underwater Wireless Sensor Network

S. MahmoudZadeh, D.M.W Powers, A. Atyabi, *IEEE Members*

*Abstract*—This paper describes a reflexive multilayered mission planner with a mounted energy efficient local path planner for Unmanned Underwater Vehicle's (UUV) navigation throughout complex subsea volume in a time variant semi-dynamic operation network. The UUV routing protocol in Underwater Wireless Sensor Network (UNSW) is generalized with a homogeneous Dynamic Knapsack-Traveler Salesman Problem emerging with an adaptive path planning mechanism to address UUV's long-duration missions on dynamically changing subsea volume. The framework includes a base layer of global path planning, an inner layer of local path planning and an environmental sub-layer. Such a multilayer integrated structure facilitates the framework to adopt any algorithm with real-time performance. The evolutionary technique known as Differential Evolution algorithm is employed by both base and inner layers to examine the performance of the framework in efficient mission timing and its resilience against the environmental disturbances. Relying on reactive nature of the framework and fast computational performance of the DE algorithm, the simulations show promising results and this new framework guarantees a safe and efficient deployment in a turbulent uncertain marine environment passing through a proper sequence of stations considering various constraint in a complex environment.

## I. INTRODUCTION

UNMANNED Underwater Vehicles (UUVs) are useful and cost effective technologies that can handle long range missions and carry out various tasks. UUVs are capable of operating close to seabed, which makes them appropriate tools for offshore industries and other scientific purposes. Enhancing the levels of motion autonomy improves vehicle's long-range operations with minimum supervision in severe and unknown environments. Motion planning significantly impacts efficiency of UUV's operation, which is the process to determine the efficient path to safely guide the UUV in critical marine missions. A sufficient level of autonomy is required to guarantee a reliable deployment and a successful mission [1]. The main concerns in UUVs' motion planning are the path reliability, path efficiency in terms of time, energy, and also uncertainties in the environment. Depending on the purpose of a mission, some other factors can also encountered such as complementing a specified number of tasks or visiting a sequence of stations which some of them can be mobile or uncertain in position (e.g. buoyant wireless sensors).

Somaiyeh MahmoudZadeh, Faculty of Information Technology, Monash University, Australia (somaiyeh.mahmoudzadeh@monash.edu)
David M.W Powers, School of Computer Science, Engineering and Mathematics, Flinders University, Australia (david.powers@flinders.edu.au)
Adham Atyabi Seattle Children's Research Institute, University of Washington, Washington, USA (adham.atyabi@seattlechildrens.org)

A great attention has been consecrated in past years on UUVs optimal motion planning to autonomously handle longer missions while reducing the operation cost [2, 3]. Thorough investigations on capability and efficiency of utilizing the Inverse Dynamics in the Virtual Domain (IDVD), a pseudo spectral method was proposed to provide real-time updates of feasible trajectory for a single Autonomous Underwater Vehicle (AUV) [4, 5]. In general, the applied motion planning algorithms for UUV's can be categorised in two disciplines of pre-generative approach [6], which is not the case of this study; and the reactive and online path planning approaches that is investigated in past few years [7]. In online path planning approaches, computational cost and the real-time performance are of extreme importance [8, 9].

Evolutionary Algorithms (EA) are proven to be competitive to traditional and well-known path planning methods (especially in applications that require real-time adjustment to the path and contain unknown trains) with occasional ability to generate near optimal solutions [9, 10]. EA are cost based non-deterministic optimization methods that emulate natural phenomenon to efficiently solve complex problems. Alvarez et al., [11] utilized Genetic Algorithm (GA) with a grid partitioning method to determine an energy efficient path for an AUV encountering strong time varying oceanic current. Zhu and Luo [12] used an improved GA to address the shortest path problem as a combinatorial optimization problem, where the Dijkstra algorithm is applied as a local search operator to generate local shortest path trees. Roberge et al., [10] developed an energy efficient path planner using meta-heuristic search nature of the Particle Swarm Optimization (PSO) algorithm, where time-varying oceanic current is encountered in the path planning process. They applied a parallel programming paradigm of "multiple-data single-program" to control the execution time. A large-scale AUV route planning has been investigated in [13] by transforming the problem space into a NP-hard graph context. The heuristic search nature of PSO and GA are employed to find the efficient sequence waypoints. Although suggested evolution-based approaches fairly satisfy the path planning constraints for autonomous vehicles, the UUV-oriented motion planning still require improvements.

In the proceeding study, the motion planning of UUV in a dynamic waypoint cluttered environment is defined and modeled as a particular type of the Dynamic Knapsack and Travelling Salesman Problems (DKTSP), where the UUV is obligated to meet a series of wireless sensors deployed at the dynamic underwater operation field. The vehicle should autonomously maximize the efficiency of the mission while dealing with changes of the environment. For achieving a maximum efficiency in a single mission, the number of visited wireless sensors and predefined stations should be maximized.

This means maximizing the use of battery capacity in the mission time instead of minimizing the travelled distance which is the case with the existing classical TSP solutions. For this purpose, a dynamic multilayer structure is developed to accommodate objectives of a UUV with a limited energy source. The system solution will require multiple actions in long range operations, which includes a continuous monitoring of the environmental status, dealing with sudden changes, dynamic reallocating the orders of stations to be visited, and managing the existing time/battery. The underwater environment is complex, dynamic, and uncertain. Considering such environmental impacts on vehicle's motion planning is highly important; however, the majority of the aforementioned studies address only a part of this impact.

*A. Research Contribution*

To address UUV's large-scale motion planning, this paper contributes a multilayered framework including a base layer for global path planning in a semi dynamic UWSN; an inner layer for local path planning in a turbulent uncertain volume, and an environmental sub-layer to model operation field and environmental influence factors. The base layer, in this context, produces a general path by appropriate prioritizing the order of buoyant-fixed stations (sensors) and guarantees on-time arrival to the destination considering deformation of the network over the time. The inner layer produces a safe path (with minimum time/energy cost) among pairs of sensors and concurrently refine the path according to environmental changes, which guarantees a safe deployment in a turbulent uncertain marine environment. The base layer splits the operating field to smaller windows (between pair of sensors). This considerably accelerates the re-planning process as less data is required to be rendered, analyzed, and recomputed.

A unique feature of this reactive framework is that both layers operate concurrently and exclusively while back feeding each other and sharing updates in both large and small scales. Large-scale update includes deformation of the operating network and orientation of the floating sensors at time and space; and small-scale update includes higher resolution information's about behavior of uncertain mobile obstacles and water current variations over the time. The parallel and concordant operation of these two layers accelerates the computation process. The efficiency of this framework is due to its multilayer integrated structure, which makes it flexible in adopting any algorithm with real-time performance. Differential Evolution (DE) algorithm is employed by both base and inner layers of the proposed model to handle UUV time restricted motion planning in a semi-dynamic operation network in the presence of environmental disturbances. Relying on reactive nature of the framework and fast computational performance of the DE algorithm, the simulations show promising results.

*B. Research Assumptions and Problem Statement*

The vehicle experiences both fixed and uncertain moving stations, in which the ocean current imposes changes to the known position of some stations (moving stations) in a bounded zone. The moving stations are the buyout moored wireless sensors and the vehicle can update its mission by reaching to one of these stations. Moreover, a clustered real map data, dynamic obstacles, and ocean time varying current maps are conducted to emulate a realistic marine environment. Assuming that the stations and buoyant wireless sensors are distributed erratically in the operation field, the following critical problems need to be addressed:

− The UUV should visit the maximum possible sensors in its path toward the goal point and dynamically compromise among order/importance of the nodes and existing battery/time (analogous to DKTSP problem).
− The inner layer is responsible to direct vehicle in smaller operation windows between pairs of stations until the vehicle reaches to the destination(s) in the network. It should provide safe and efficient maneuvers, deal with dynamicity of the ocean, and reduce the path time, which is another NP-Hard problem.
− Current force and its relative orientation can strongly affect displacement of the vehicles. The adverse current can deflect vehicle from generated trajectory, so it should be avoided. Accordingly, the accordant current can be useful for saving energy, but it may take longer distance. The vehicle should be able to understand the circumstances and trade-off between the optimisation criteria, such as path reliability, time/battery cost, and distance.

The environmental influences along with vehicle's kinematic properties should be taken into the account in optimal path generating process. To this purpose, a sub-layer is defined to provide a pre-analysis process for modelling the environmental framework.

The rest of this paper is organised as follows; Section 2 describes the mathematical modelling of the underwater operation field, operational requirements, and the distributed floating wireless sensor network. Section 3 introduces the multilayered UUV motion planning framework and provides the problem formulation according to the complex DKTSP problem. An overview of the DE-based navigation and its application on existing approach is provided by Section 4. The simulation results of the integrated multilayered approach are also discussed in this section. Section 6 concludes the paper.

## II. MODELLING THE OPERATION FIELD AND OPERATIONAL REQUIREMENTS

A UUV operates in a highly variable and uncertain environment, which this variability can have devastating effect on mission timing and vehicle's motion. Hence, a proper modelling of the environment and applying necessary assumptions play a critical role in performance of the motion planning. The engaged environmental factors in this research are modelled in the following subsections.

*A. Underwater Wireless Sensor Network (UWSN) Integration and Eligible Operation Area*

Interfacing with static-mobile sensor network systems is the highlighted particularity of majority of the robotic systems at Commonwealth Scientific and Industrial Research Organisation (CSIRO) [14]. The base layer receives the topology of UWSN, position of stations (sensors), and adjacency information of the network as input. In UWSNs, sensor nodes are for data delivering and presented as stations for the UUV. The vehicle moves through the stations following the path generated by the base layer, where on-time visit to the target station is the main concern of this layer. The UUV broadcast its position in the map to the closest sink, as they get visited [15]. By doing so, the other sensors associate their regions with the UUV. The vehicle moves along the sensors in a way to ensure the maximum data collection [16]. A sample of such a network is depicted by Fig.1.

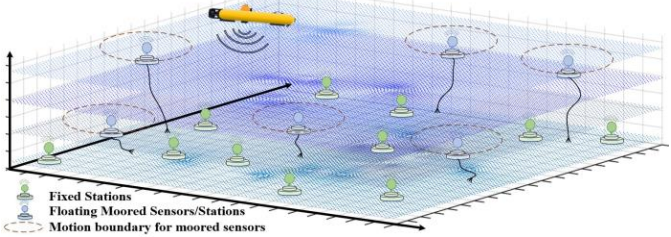

Fig. 1. A sample of UWSN representation in the operating area.

Existence of prior information about the environment, geographical position of stations, and location of coasts as the forbidden zones for deployment has a significant impact on UUV's efficient and robust motion planning. In order to have a realistic model of the operation area, a sample geographic map is utilized in this study. The coastal zone and the authorized water-covered zone are separated by K-means clustering method [8, 17], where map data get partitioned into $k$ groups of $M=\{M_1,…, M_k\}$ and the algorithm is initialized with $k$ cluster centers. The clusters get refined iteratively until the saturation phase emerges. The objective function here is a squared error function in (1) that should be minimized.

$$\begin{cases} \partial:\{\partial_1,...,\partial_n\} \\ M:\{M_1,...,M_k\} \end{cases} \Rightarrow \arg\min_M \sum_{i=1}^{k} \sum_{\partial \in M_i} \|\partial - \mu_i\| \quad (1)$$

Here, the $\partial$ represents a set of observations, and $\mu$ represents the cluster centers of $M=\{M_1,…, M_k\}$. The algorithm uses the geographical map in form of an image of size 1000-by-1000 pixels where each pixel represents $1\times1$ m$^2$. Water-covered zones are recognized by k-means as the eligible sections for UUV operation. The network topology of the UWSN is also configured and deformed in the water-covered area. The base and inner layer planners operate in the water-covered feasible regions. To fulfil the operational requirements, the UUV is equipped with a set of navigation sensors, such as sonar sensors, compass, depth/pressure sensor, and camera. A set of fixed and floating moored stations are defined in the network where the moored stations/sensors are modelled according to the FleckTM wireless sensor network devices [14]. Stations are distributed across the operating area throughout the water column. Efficient visiting to the sensors facilitates the vehicle to early detection of events. Therefore, the UWSN associated with two types of stations as follows:

a) The fixed stations in the 3dimensional volume represented by $fS_{xyz}$ with known locations initialized once in advance by a uniform distribution of

$$\forall fS_{xyz} \Rightarrow \begin{cases} fS_x \sim U(0,10000) \\ fS_y \sim U(0,10000) \\ fS_z \sim U(0,1000) \end{cases} \quad (2)$$

b) Dynamic moored sensors presented by $dS_{xyz}$ that feed information of local area to the vehicle. These sensors broadcast their locations. These information is sensed by the UUV's sonar sensors with a specific uncertainty modelled by (3). These sensors move along a direction motivated by current velocity. Hence, the $dS_{xyz}$ position has a truncated normal distribution, where its probability density function defined by (4), as follows:

$$\begin{aligned} & Bound_{xyz}^{\max} \sim N(0,\sigma^2) \\ & |dS_{xyz}| \leq Bound_{xyz}^{\max} \\ & U_R^C = |\upsilon_c| \sim N(0,0.3) \\ & f\left(dS^i;0,\sigma^2,Bound_{xyz}^{\max}\right) = \frac{dS_{xyz}^i + U_R^C}{|2 \times Bound_{xyz}^{\max}|} \\ & dS_{xyz}^i \ \& \ fS_{xyz}^i \notin \{MAP=0\} \end{aligned} \quad (3),(4)$$

the $U_R^C$ corresponds to uncertainty on sensors motion which is considered equal to magnitude of the water current $|\upsilon_c|$. The {MAP=0} represents the coastal areas in the clustered geographical map.

Due to energy restrictions, a single UUV cannot visit all stations in a single mission in a large-scale operation field. Therefore, the order of stations should be prioritized based on some influence factors that is used by the base layer of the framework. This can provide a priority tour to have a beneficial mission, while reaching to the goal station is the most critical priority for the vehicle. Hence, the stations should be prioritized and selected in a way that govern the UUV to the destination.

*B. Ocean Current and Uncertain Fixed-Mobile Obstacles as Environmental Influence Factors*

Several factors can influence the underwater current and impose operational constrains. For example, the ocean currents may have positive or negative effect on vehicles motion. Hence, the UUV's attitude and position should be adjusted constantly according to water current variations. The environmental model constructed in this study considers both water current and fixed-mobile uncertain objects. This study takes the use of a numerical estimate model derived from Navier-Stokes equations and Lamb vortices [18] to model 2D turbulent current map. Hence, a 2D pattern is applied as the update of deep ocean current fields is very slow and inappreciable, which is negligible in large-scale operations. Moreover, the UUV's motion usually is assumed on horizontal plane, as its vertical motions is comparatively negligible due to large horizontal scales [19]. The UUV uses the following physical model to diagnose the current fields' orientation:

$$\begin{aligned} & \vec{\upsilon_c}:(\upsilon_{c,x},\upsilon_{c,y}) = f(\vec{\Gamma}^O, \Im, \ell) \\ & \upsilon_{c,x}(\vec{\Gamma}) = \left(\Im e^{-(\vec{\Gamma}-\vec{\Gamma}^O)^2 \ell^{-2}} - \Im\right)(y - y_0) \Big/ 2\pi(\vec{\Gamma}-\vec{\Gamma}^O)^2 \\ & \upsilon_{c,y}(\vec{\Gamma}) = \left(\Im - \Im e^{-(\vec{\Gamma}-\vec{\Gamma}^O)^2 \ell^{-2}}\right)(x - x_0) \Big/ 2\pi(\vec{\Gamma}-\vec{\Gamma}^O)^2 \end{aligned} \quad (5)$$

here, $\Gamma$ is a 2D volume. The $\Gamma^o$, $\ell$ and $\Im$ are vortex center, radius, and strength, respectively. $\upsilon_{c,x}$ and $\upsilon_{c,y}$ are the horizontal components of the current velocity vector $\upsilon_c$. Coastlines and islands are considered as static obstacles, where their locations are known in advance from the offline map. Some uncertain areas are considered as forbidden flying zones and modelled as uncertain obstacles with varying radius in a specified bound of $\Theta_r = \mathbf{N} \sim (\Theta_p, \sigma_0)$, where the $\Theta_r$ and $\Theta_p$ correspond to obstacles radius and position, respectively. The $\Theta_r$ changes iteratively. The vehicle also experiences mobile obstacles in its deployments. These objects may be influenced by current flow or may move spontaneously. The current effect on these objects are modeled by a multidirectional uncertainty propagation, proportional to current magnitude $U_R^C = |v_c| \sim (0,0.3)$. Obstacles motion and coordinates can be detected by sonar sensors using (6):

$$\Theta_p(t) = \Theta_p(t-1) \pm U(\Theta_{p_0}, \sigma)$$
$$\Theta_r(t) = \Theta_r(t-1) \begin{bmatrix} 1 & U_R^C(t) & 0 \\ 0 & 1 & 0 \\ 0 & 0 & 1 \end{bmatrix} + N(0,\sigma) \begin{bmatrix} 0 \\ 1 \\ 1 \end{bmatrix} + \begin{bmatrix} 0 \\ 0 \\ U_R^C(t) \end{bmatrix} \quad (6)$$

## III. MULTILAYERED MOTION PLANNING FRAMEWORK

This study aims to develop a reactive multilayered motion planning framework to increase a UUV efficiency in long-range missions considering severe underwater environmental influences. The framework accommodates two reactive and deliberative execution layers, which is depicted by Fig.2. The reactive layer, called the "base layer", is designed to maximize mission efficiency by appropriate ordering the stations while managing mission time and directing the vehicle toward the goal station.

The UWSN deforms over the time due to displacement of the floating sensors/stations and these sensors continuously broadcast their location. Location of the buoyant sensors in the network is assumed to be changed in a bounded zone with a specific uncertainty (see (3) and (4)). Accordingly, the base layer simultaneously rearranges the order of stations to be visited, taking passage of time into account. The deliberative layer is called the "inner layer", deals with local variation of the environment and aims to perform prompt reactions to sudden events. This layer comprises automatic functions operated in background to boost vehicle's self-supervision characteristics. The base layer splits the operation area to smaller sections (bounded to the distance between pairs of stations) for the inner layer; reciprocally the inner layer produces safest and quickest path along the ordered stations, while deals with continuous variation of the sub-area in proximity of the vehicle. A sub-layer of "environment layer" is also designed to provide required information for the base and inner layers. Another operation handler is embedded to the base layer to synchronize operation of different layers and to reclaim the probable missed time in their parallel execution.

### A. Motion Planning as the Dynamic Knapsack and Travelling Salesman Problem (DKTSP)

As mentioned previously, the base layer aims to find the most efficient path which includes finding a beneficent sequence of stations while optimizing some performance factors such as time and distance. This problem can be considered as assimilation of the Dynamic Knapsack and Travelling Salesman Problem (DKTSP) [20,21], in which there should be an equipoise among time/battery, distance, serving maximum possible stations, ensuring on-time arrival to the goal station, before the UUV runs out of battery. Many recent studies have considered TSP for simplifying their path-route planning problem. Dunbabin et al., [14] used TSP for modelling AUV routing in an UWSN. In their study, AUV been used to collect environmental data over the network. However, optimal path was not guaranteed as the uncertainty and variability of the environment was not considered. UAVs path planning for information collection is another relevant research carried out by Ergezer and Leblebicioglu [21] that used TSP for modelling their graph routing problem assuming a static environment.

*Knapsack* is a decision problem by the content of achieving the maximum value in item selection but not exceeding a specific weight (knapsack's capacity), which is a Non-Polynomial (NP) complete problem [22]. On the other hand, *TSP* is another NP-complete problem that aims to determine a best and efficient route in a given list of *m* cities, while minimising the travelling cost [23]. TSP has different versions based on the defined constraints. For instance, in a typical version of TSP, cities should be visited only once. In our problem, this constraint is set on edges rather than nodes. Finding a solution in a large number of possible solutions that is equal to "(*m*-1)!/2", makes the TSP complex and challenging. Increase in the size of network (number of nodes/connections) makes this problem even more challenging. Considering a TSP problem configuration for our study, stations can represent cities to be visited in a way that a certain sequence of stations is to be visit before arriving to destination. Incorporating Dynamic Knapsack (DK) configuration to our study, stations are prioritized, while their priority and distribution is changing over the tame and space. Accordingly, estimated time for visiting the set of selected stations should be fitted to the time constraint of the problems. In this study, the UWSN's topology is changing

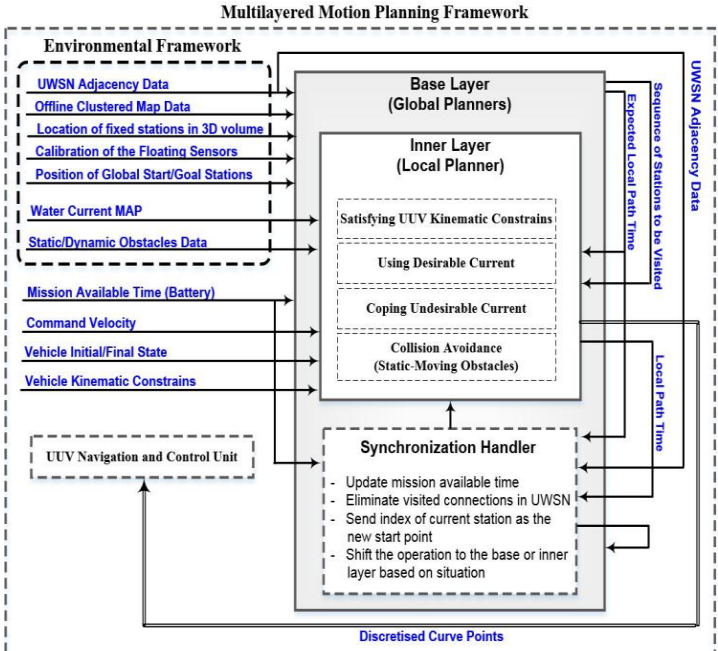
Fig.2. The operation diagram of the introduced reactive multilayered framework.

over the time. Indeed, in this particular case unlike a typical TSP, maximizing the distance travelled would be more favourable due to serving more stations (collecting more data). These problems can be represented in different ways. Here, we represent it as a permutation problem, where edges have weight and cost.

$$\forall\, dS_{xyz}^i\; \&\; fS_{xyz}^i \notin \{MAP=0\};\quad \{S_{xyz}\}=\{dS_{xyz}\}\cup\{fS_{xyz}\}$$
$$\Re_k = \left(S_{xyz}^1,...,S_{xyz}^i,...,S_{xyz}^n\right) \quad (7)$$

Where, $S_{xyz}^i$ represents the coordinate of any arbitrary station in geographical frame (which is a collection fixed stations $fS_{xyz}$ and dynamic stations $dS_{xyz}$). $\Re$ is an arbitrary sequence of nodes. In UWSN, any arbitrary edge $E_{ij}$ is characterized by the corresponding distance $D_{ij}$, required traversing time $T_{ij}$, which is calculated by (8):

$$\forall\, dS_{xyz}^i\; \&\; fS_{xyz}^i;\quad \exists\, E_{ij}:(D_{ij}(t),T_{ij},\rho_{ij})$$
$$D_{ij}(t)=\sqrt{\left(S_x^j(t)-S_x^i(t)\right)^2+\left(S_y^j(t)-S_y^i(t)\right)^2+\left(S_z^j(t)-S_z^i(t)\right)^2} \quad (8)$$
$$T_{ij}=\left(D_{ij}(t)\times|\upsilon|^{-1}\right)$$

here, the $E_{ij}$ is weighted in advance by a priority value of $\rho_{ij}$. The $|\upsilon|$ is the UUV's absolute velocity. The UWSN get altered simultaneously. Relocating of the dynamic stations cause change in distances between stations, so the $D_{ij}(t)$ is defined as a function of time, so considering network variation over the time is necessary in cost computation and re-planning process. Here, if we consider $\Pi_k$ as the collection of all permutations, the DKTSP is the searching of $\Re \in \{\Re_1,\ldots,\Re_k\}$ in $\Pi_k$ such that

$$\Re_r = \left(S_{xyz}^1(t),...,S_{xyz}^i(t),...,S_{xyz}^n(t)\right)$$
$$\vec{E}_{ij}:\left\langle S_{xyz}^j(t)-S_{xyz}^i(t)\right\rangle$$
$$D_\Re(t)=\sum_{\substack{i=0\\j\neq i}}^{n} lE_{ij}\times\sqrt{\left(S_{xyz}^j(t)-S_{xyz}^i(t)\right)^2};\quad l\in\{0,1\} \quad (9)$$
$$T_\Re = \left(D_\Re(t)\times|\upsilon(t)|^{-1}\right)$$

where, $D_\Re$ and $T_\Re$ are the route ($\Re_r$) distance and time, respectively. The $T_\Re$ should approach total available time $T_M$ but not overstep it. The total value of route ($\rho$) should be maximized. Hence route cost of $C_\Re$ is calculated as follows:

$$C_\Re \propto \sum_{\substack{i=0\\j\neq i}}^{n}\left|lE_{ij}D_{ij}(t)\times|\upsilon|^{-1}-T_M\right| = \sum_{\substack{i=0\\j\neq i}}^{n}\left|T_{\Re r}-T_M\right|+\left(lE_{ij}\times\rho_{ij}\right)^{-1}$$
$$s.t.\quad \forall\, \Re_r;\quad \max(T_{\Re r})<T_M \quad (10)$$

The $C_\Re$ has a direct relation to the passing distance among stations ($D_\Re$). On the other hand, the inner layer is responsible to find the collision-free safest and shortest path between two stations, while take the environmental dynamics into account. The criteria for the inner layer is minimizing the distance $D_{ij}$, and time $T_{ij}$, dealing with water current variations, and avoiding collision. Adverse current can deviate vehicle from its trajectory and cause extra battery consumption. In contrast, accordant current can motivate its motion and cause energy saving. Hence, current influence on UUV's motion and kinematics should be investigated. UUV's six degree of freedom in translational and rotational motion in NED $\{n\}$ and Body $\{b\}$ frame is given by (11) [24]. The effect of current flow on vehicle's motion is determined by (12) as follows:

$$\{b\}\rightarrow \eta:\langle X,Y,Z,\varphi,\theta,\psi\rangle$$
$$\{n\}\rightarrow \upsilon:\langle \upsilon_x,\upsilon_y,\upsilon_z,p,q,r\rangle \quad (11)$$

$$\vec{\upsilon}_c=[\upsilon_{c,x},\upsilon_{c,y}]\Rightarrow \begin{cases}\upsilon_{c,x}=|\upsilon_c|\cos\theta_c\cos\psi_c\\ \upsilon_{c,y}=|\upsilon_c|\cos\theta_c\sin\psi_c\end{cases}$$
$$\upsilon_x=|\upsilon|\cos\theta\cos\psi+|\upsilon_c|\cos\theta_c\cos\psi_c \quad (12)$$
$$\upsilon_y=|\upsilon|\cos\theta\sin\psi+|\upsilon_c|\cos\theta_c\sin\psi_c$$
$$\upsilon_z=|\upsilon|\sin\theta$$

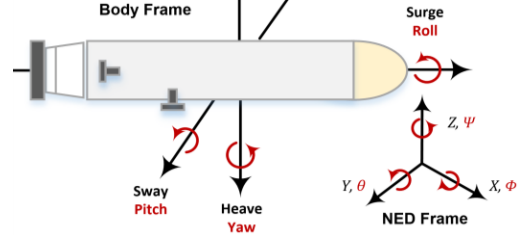

Fig. 3. The UUV's kinematic in a 3D volume.

here, the UUV's North ($x$), East ($y$), and Depth ($z$) position along the path is presented by $X$, $Y$, and $Z$, respectively. The $\varphi$, $\theta$, $\psi$ are the UUV's Euler angles of roll, pitch, and yaw. The $\upsilon_x$, $\upsilon_y$, $\upsilon_z$ are surge, sway and heave water referenced velocities of the UUV's in $\{b\}$-frame. The potential path corresponding to distance $D_{ij}$ is generated based on B-Spline curves [25, 26] captured from a set of control points $\vartheta$ as follows:

$$\vec{E}_{ij}:\left\langle S_{xyz}^j-S_{xyz}^i\right\rangle;\quad S_{xyz}^i\equiv \vartheta_{xyz}^1;\quad S_{xyz}^j\equiv \vartheta_{xyz}^m$$
$$\vartheta:\{\vartheta_{xyz}^1,...,\vartheta_{xyz}^k,...,\vartheta_{xyz}^m\} \quad (13)$$
$$\vartheta_{xyz}^k(t)=L_{\vartheta,xyz}^k+Rand_k\left(U_{\vartheta,xyz}^k-L_{\vartheta,xyz}^k\right)$$
$$\psi(t)=\arctan\left(\left|\left(\Delta\vartheta_y^k(t)\right)^2\right|\left|\left(\Delta\vartheta_x^k(t)\right)^2\right|^{-1}\right)$$
$$\theta(t)=\arctan\left(\left(-\left|\left(\Delta\vartheta_z^k(t)\right)^2\right|\right)\left(\left(\Delta\vartheta_y^k(t)\right)^2+\left(\Delta\vartheta_x^k(t)\right)^2\right)^{\frac{-1}{2}}\right)$$
$$X(t),Y(t),Z(t)=\sum_{k=1}^{m}B_{k,K}\times\left\{\vartheta_x^k(t),\vartheta_y^k(t),\vartheta_z^k(t)\right\} \quad (14)$$
$$D_{ij}(t)\equiv \wp_{xyz}^{ij}=\sum_{k=1}^{m}\sqrt{(\Delta\vartheta_x^k(t))^2+(\Delta\vartheta_y^k(t))^2+(\Delta\vartheta_z^k(t))^2}$$
$$\wp(t)=[X(t),Y(t),Z(t),\psi(t),\theta(t),\upsilon_x(t),\upsilon_y(t),\upsilon_z(t)]$$

where, $B_{k,K}$ is the curve's blending functions, and $K$ is its smoothness factor. Control points are located in bundle of $\vartheta \in [U_\vartheta^k, L_\vartheta^k]$ in Cartesian coordinates, where the $U_\vartheta^k, L_\vartheta^k$ are the predefined upper and lower bounds. Appropriate adjustment of the control points plays a substantial role in optimality of the generated path $\wp_{xyz}^{ij}$ along the $D_{ij}$.

### B. Multilayered Motion Planning Optimization Criterion

The total route cost $C_\Re$ is directly linked to distance between each pair of stations $D_{ij}$; hence, the local path cost $C_\wp$ for any optimum local path gets used in the context of the $C_\Re$. The local path generated by the inner layer gets evaluated by the path time (proportional to energy consumption and travelled distance) and its collision avoidance capability. It should be feasible and meet the environmental and vehicle kinematic constraints (described by (15)). The environmental constraints are associated with coastal area of map and fixed-mobile obstacles; and the kinematic constraints associated with UUV's yaw surge and sway rates. Therefore, the final solution generated by framework gets evaluated as follows:

$$\forall \wp_{xyz}^{ij} \Rightarrow \begin{cases} D_{ij} \equiv \wp_{xyz}^{ij}(t) = \sum_{\substack{k=1 \\ \vartheta^1 \equiv i, \vartheta^m \equiv j}}^{m} \left|\vartheta_{xyz}^{k+1}(t) - \vartheta_{xyz}^{k}(t)\right| \\ T_{ij} \equiv T_\wp = D_{ij}(t) \times |\upsilon(t)|^{-1} \end{cases}$$

$$\Sigma_{M,\Theta} = \begin{cases} 1 & \wp_{xyz}^{ij}(t) \cap Coast : MAP(x,y) = 1 \\ 1 & \wp_{xyz}^{ij}(t) \cap \bigcup_{N\Theta} \Theta(\Theta_p, \Theta_r) \\ 0 & Otherwise \end{cases} \quad (15)$$

$$C_\wp = T_\wp + \varepsilon_1 \max(0; \upsilon_x(t) - \upsilon_{x,\max}) + \varepsilon_2 \max(0; |\upsilon_y(t)| - \upsilon_{y,\max}) + ...$$
$$... + \varepsilon_3 \max(0; |\dot\psi_t| - \psi_{\max}) + \varepsilon_4 \Sigma_{M,\Theta}$$

$$D_{ij} \equiv \wp_{xyz}^{ij}; \quad T_{ij} \equiv C_\wp^{ij}; \quad l = \{0,1\}$$

$$T_\Re \propto \sum_{\substack{i=0 \\ j \neq i}}^{n} lE_{ij} \times \wp_{xyz}^{ij}(t) \times |\upsilon|^{-1} \Rightarrow T_\Re = \sum_{\substack{i=0 \\ j \neq i}}^{n} lE_{ij} \times C_\wp^{ij} \quad (16)$$

$$C_\Re = \left|\sum_{\substack{i=0 \\ j \neq i}}^{n} lE_{ij} \times C_\wp^{ij} - T_M\right| + (lE_{ij} \times \rho_{ij})^{-1} + \max(0, T_\Re / T_M)$$

$\sum_{M,\Theta}$ is the collision violation function, and $\varepsilon_1$, $\varepsilon_2$, $\varepsilon_3$, and $\varepsilon_4$ represent the impact of each constraint violation in local path cost $C_\wp$ calculation. Number of stations is denoted by $n$, and $l$ is a selection variable.

*C. Global-Local Motion Re-planning*

The base layer starts its operation by generating an initial optimum global path (sequence of stations) according to primary topology of the NWSN. Then the inner layer starts its operation by generating local optimum path between pairs of stations listed in the global path, while handling dynamicity of the environment. The inner layer also applies on-line re-planning for correcting the previous trajectory whenever it faces a disturbance. Accordingly, the environmental situation is monitored simultaneously, and the local path is refined continually from the existing position to target point. As the inner layer incorporate any dynamic changes of the environment, its operation may take longer than expected, which means the local path time $T_\wp^{ij}$ exceeds the expected time $T_{ij}$ for passing the distance $D_{ij}$. In this case, a part of available time $T_M$ is wasted, and the previously generated global path cannot be optimum anymore. On the other hand, the UWSN's topology may change due to buoyant sensors. Some sensors also may lose their energy and leave the network. Under these circumstances, the base layer should re-plan a new global path considering the present topology of the UWSN. In this process, after visiting each station, the $T_M$ gets updated; the passed edge(s) get eliminated from the UWSN; update of the UWSN topology is fed to the base layer, and the position of the current station is sent to both base and inner layers as the new starting point. Afterward, the base layer generates a new global path according to network updates. This process continues until UUV arrives to the goal station. The Fig.4 clarifies the local and global re-planning process.

With respect to above discussion, the framework should be fast enough to monitor the environmental updates, handle the changes, and provide prompt local-global re-planning according to situation. This complex process is not a trivial problem and brute force algorithms cannot be sufficient for quick computation of the optimal solutions [27]. On the other hand, for the UUV large-scale path-planning, having a quick acceptable path that satisfies all constraints is more appropriate rather than taking a long computational time to find the best path. Currently, no polynomial time algorithm is known yet to solve such a complex NP-hard problem [24]. Instead, it has been shown by many studies that meta-heuristics are viable candidates to effectively solve such a complex problem within an affordable time [10, 28]. In this work, DE is employed to produce the local-global path for a UUV in an uncertain maritime environment.

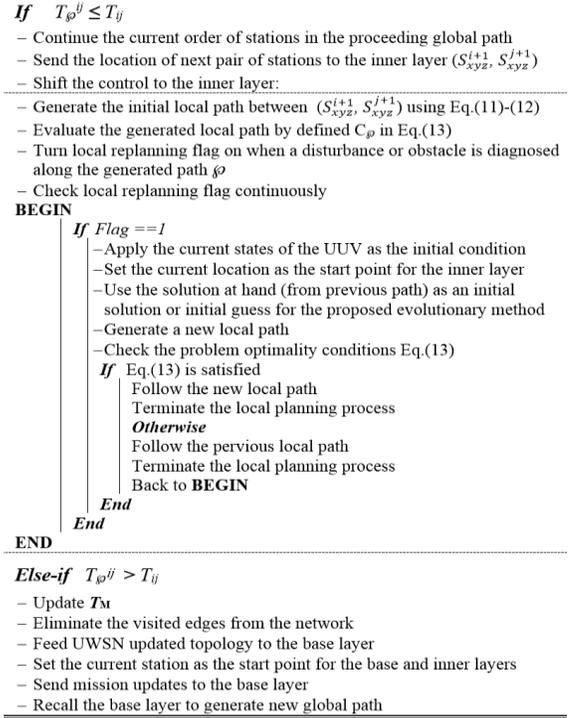

Fig.4. Reactive local and global re-planning mechanism.

## IV. DIFFERENTIAL EVOLUTION ON MULTILAYERED PATH PLANNING PROBLEM

Several extensions to evolutionary methods have been suggested in recent years offering improved performance on benchmark problems and DE has shown superior performance in several real-world applications [29-31]. The DE is a cost based non-deterministic optimization method that follows evolution-inspired search method to efficiently solve complex problems. The DE is applicable on different problems since the generated solutions satisfy the algorithms basic requirements and the defined objective function. Vesterstrom and Thomsen [30] evaluated the performance of DE comparing to some EA approaches such as PSO and GA through the 34 widely used benchmark problems.

The DE algorithm is an improved version of GA and applies similar evolution operators but uses floating-point numbers and real coding for presenting problem parameters that enhances solution quality and provides faster optimization [29]. The structure of algorithm is mostly based on non-uniform crossover and differential mutation operations. The DE exposes flexible mechanism in terms of operators' construction and parameters penalty-tuning. It uses crossover operator to shuffle the existing population, then a selection operator is applied to approach the solutions to the desirable solution space. Thus, with the clarity gained by hindsight from those works, the DE algorithm is employed and tuned for the proposed UUV motion planning problem in the paper. We seek to apply the most effective construction of the operators to fit the algorithm suitable for UUV's local-global operation considering a dynamic ocean structure and recalibration of the floating moored sensor nodes.

### DE-Based Global Path Generation (in the base layer)

Suitable vector coding is the most critical step of DE process that directly impacts the algorithm's performance. This global path planner deals with finding the optimal path through the UWSN. Considering the UWSN topology, a solution vector is feasible when it starts with the position of the start station and ends with the position of the target station. The vector should not include extinct or repeated connections in the network. To keep the solutions feasible, a priority vector is applied to the initialization phase of the algorithm, in which the station sequence in each vector is selected according to their priority values and UWSN adjacency guiding information. More details on this process can be fined in a former research [9]. Mutation in DE is the main wheel behind its impressive performance that uses a weighted difference vector between two population members to a third one (is called *donor*). A proper donor accelerates convergence rate. These three vectors of $\chi_{r_1,t}$, $\chi_{r_2,t}$, and $\chi_{r_3,t}$ are selected randomly from the same generation $t$. So, the mutant solution vector is produced by

$$\dot{\chi}_{i,t} = \chi_{r_3,t} + S_f(\chi_{r_1,t} - \chi_{r_2,t})$$
$$r_1 \neq r_2 \neq r_3 \neq i \in \{1,...,i_{max}\}, \quad S_f \in [0,1+] \quad (17)$$

where, $S_f$ is a scaling factor that controls the amplification of the difference vector. Giving a proper value to $S_f$ improves the exploration capability of DE. Having a proper *donor* accelerates convergence rate. In this approach, the *donor* is determined randomly with uniform distribution as follows:

$$donor = \sum_{i=1}^{3}\left(\lambda_i \left(\sum_{j=1}^{3} \lambda_j\right)^{-1}\right) \times \chi_{ri,t}, \quad (18)$$

The $\lambda_j \in [0,1]$ is a uniformly distributed value. The mutant and parent vectors $\dot{\chi}_{i,t}$ and $\chi_{i,t}$ are then shifted to the crossover operation as follows:

$$\begin{aligned}\chi_{i,t} &= (x_{1,i,t},...,x_{n,i,t}) \\ \dot{\chi}_{i,t} &= (\dot{x}_{1,i,t},...,\dot{x}_{n,i,t}) \\ \ddot{\chi}_{i,t} &= (\ddot{x}_{1,i,t},...,\ddot{x}_{n,i,t})\end{aligned} \Rightarrow \ddot{x}_{j,i,t} = \begin{cases} \dot{x}_{j,i,t} & rand_j \leq C_r \vee j = k \\ x_{j,i,t} & rand_j \leq C_r \wedge j \neq k \end{cases}$$
$$j = 1,...,n; \quad n,k \in [1,i_{max}] \quad (19)$$

where, $\ddot{\chi}_{i,t}$ is the produced offspring at iteration $t$; $k$ is a random index chosen once for all population; $C_r \in [0,1]$ is the crossover factor, set by operator. The offspring produced by the crossover and mutation operations is evaluated using defined cost function, then offspring with the minimum cost are transferred to the next generation and the rest will be eliminated. Solutions get improved at each iteration. The DE process of global path planning is given by Fig.5.

### DE-Based Local Path Generation (in the inner layer)

In the local path planning, an initial population of solution vectors $\chi_i^{\wp}$, $(i=1,...,i_{max})$ is randomly generated with uniform probability, where any arbitrary path $\wp$ is assigned with solution vector $\chi_i^{x,y,z}$ in a 3D volume, in which control points $\vartheta$ along the path $\wp$ correspond to elements of the $\chi_i^{x,y,z}$. The solution space efficiently gets improved iteratively applying evolution operators. A candidate solution vector for a local path is designated as

$$\forall \chi_{i,t}^{\wp} : (\chi_{i,t}^x, \chi_{i,t}^y, \chi_{i,t}^z) \atop i \in \{1,...,i_{max}\} \atop t \in \{1,...,t_{max}\} \Rightarrow \begin{cases} \chi_{i,t}^x = \vartheta_x^i \\ \chi_{i,t}^y = \vartheta_y^i \\ \chi_{i,t}^z = \vartheta_z^i \end{cases} \quad (20)$$

After primary population initialized, the DE process is similarly repeated for local path planning as represented by Fig.6. The entire process of DE based multilayered motion planner is depicted by the given flowchart in Fig.7.

---

**DE Global Path Planning**
- Set the maximum number of iterations $t_{max}$ and population size $i_{max}$
- Initialize solution vectors randomly with uniform probability restricted to priority values and adjacency connections:
  $\chi_{i,t}^{\Re}, \quad i = \{1,...,i_{max}\} \quad t \in \{1,...,t_{max}\}$
- Set the crossover coefficient $C_r$

*For* $t = 1$ *to* $t_{max}$
  *For* $i = 1$ *to* $i_{max}$
    Reconstruct the global path $\chi_{i,t}^{\Re}$ using priority values and UWSN adjacency
    Evaluate the global path according to $C_{\Re}$ in Eq.(10).
    Determine the *donor* by Eq.(18)
    Apply mutation using $\dot{\chi}_{i,t}^{\Re} = \chi_{r_3,t}^{\Re} + S_f(\chi_{r_1,t}^{\Re} - \chi_{r_2,t}^{\Re})$
    *For* $j = 1$ *to* $i$
      Apply crossover using $\chi_{i,t}^{\Re}$ mutant solution $\dot{\chi}_{i,t}^{\Re}$ to get the $\ddot{\chi}_{i,t}^{\Re}$
      $\chi_{i,t}^{\Re} = (x_{1,i,t},...,x_{n,i,t})$
      $\dot{\chi}_{i,t}^{\Re} = (\dot{x}_{1,i,t},...,\dot{x}_{n,i,t})$ $\Rightarrow \ddot{x}_{j,i,t} = \begin{cases} \dot{x}_{j,i,t} & rand_j \leq C_r \vee j = k \\ x_{j,i,t} & rand_j \leq C_r \wedge j \neq k \end{cases}$
      $\ddot{\chi}_{i,t}^{\Re} = (\ddot{x}_{1,i,t},...,\ddot{x}_{n,i,t})$ $\quad j = [1,n]; \quad n,k \in [1,i_{max}]$
      Evaluate the $\chi_{i,t}^{\Re}$, $\dot{\chi}_{i,t}^{\Re}$, and $\ddot{\chi}_{i,t}^{\Re}$
      *if* $C_{\Re}(\chi_{i,t}^{\Re}) \leq C_{\Re}(\dot{\chi}_{i,t}^{\Re})$
        $\dot{\chi}_{i,t+1}^{\Re} = \chi_{i,t}^{\Re}$
        *if* $C_{\Re}(\chi_{i,t}^{\Re}) \leq C_{\Re}(\ddot{\chi}_{i,t}^{\Re})$
          $\ddot{\chi}_{i,t+1}^{\Re} = \chi_{i,t}^{\Re}$
        *else*
          $\ddot{\chi}_{i,t+1}^{\Re} = \ddot{\chi}_{i,t}^{\Re}$
        *end (if)*
      *else*
        $\chi_{i,t+1}^{\Re} = \ddot{\chi}_{i,t}^{\Re}$
      *end (if)*
    *end (For)*
  *end (For)*
  Select the best solutions to transfer to next iteration
*end (For)*
Output best solution and the corresponding global paths $\Re$

Fig.5. Pseudo code of DE based global path planning.

---

**DE Local Path Planning**
Initialization phase:
- Set the maximum number of iteration $t_{max}$ and the population size $i_{max}$
- Initialize population of solution vectors randomly $\chi_{i,1}^{\wp}:\chi_{i,1}^{x,y,z}$ with the control points $(\vartheta_x^i, \vartheta_y^i, \vartheta_z^i)$
- Set the crossover coefficient $C_r$

*For* $t = 1$ *to* $t_{max}$
  *For* $i = 1$ *to* $i_{max}$
    Reconstruct a path $\wp$ according to Eq.(13) and (14)
    Evaluate the local path using $C_{\wp}$ given by Eq.(15) and (16)
    Determine the *donor* by E.(18)
    Apply mutation using $\dot{\chi}_{i,t}^{\wp} = \chi_{r_3,t}^{\wp} + S_f(\chi_{r_1,t}^{\wp} - \chi_{r_2,t}^{\wp}), \quad S_f \in [0,1+]$
    Apply crossover on $\chi_{i,t}^{\wp}$, and $\dot{\chi}_{i,t}^{\wp}$ to produce $\ddot{\chi}_{i,t}^{\wp}$
    Reconstruct and evaluate path $\wp$ corresponding to $\chi_{i,t}^{\wp}, \dot{\chi}_{i,t}^{\wp}$, and $\ddot{\chi}_{i,t}^{\wp}$
    $\dot{\chi}_{i,t+1}^{\wp} = \begin{cases} \dot{\chi}_{i,t}^{\wp} & C_{\wp}(\dot{\chi}_{i,t}^{\wp}) \leq C_{\wp}(\chi_{i,t}^{\wp}) \\ \chi_{i,t}^{\wp} & C_{\wp}(\dot{\chi}_{i,t}^{\wp}) > C_{\wp}(\chi_{i,t}^{\wp}) \end{cases}; \quad \ddot{\chi}_{i,t+1}^{\wp} = \begin{cases} \ddot{\chi}_{i,t}^{\wp} & C_{\wp}(\ddot{\chi}_{i,t}^{\wp}) \leq C_{\wp}(\chi_{i,t}^{\wp}) \\ \chi_{i,t}^{\wp} & C_{\wp}(\ddot{\chi}_{i,t}^{\wp}) > C_{\wp}(\chi_{i,t}^{\wp}) \end{cases}$
  *end*
  Select the best solutions to transfer to next iteration
*end*
Output best solution and the corresponding local paths $\wp$

Fig.6. Pseudo code of DE based local path planning.

The mission at the base layer is specified as a series of stations with different functionalities such as floating sensors and fixed stations. The UUV should visit adequate number of them in a restricted time interval. The floating sensors have a changing calibration, so the topology of UWSN changes over the time. This means a sequence of produced stations may turn to be inefficient due to distance changings. The UUV attempts to pick a shortest and safest path (generated by the inner layer) between stations. The local path is refined as obstacles or disturbing current are detected. The real-time performance of the base and inner layers incorporates the deformation of network topology and facilitates the vehicle to perform prompt reaction to sudden changes of the local environment; hence, the global and local trajectory gets updated continuously up on the request. The generalized objectives of the proposed multilayered framework are:

− To produce an efficient global path fitted to the battery life-time (see (10) and (16)).
− To visit the highest value stations in the given time.
− To re-plan a new global path whenever planned time is wasted to handle any unexpected event during the mission.
− To produce a shortest local path between stations by minimizing the flight time and distance (see (15)).
− To avoid collision in local operations by refining the local path in the inner layer.
− To use accordant water current and to avoid adverse current forces in local path planning (see (15)).
− To satisfy vehicles kinematic constraints of surge-sway directional velocities and theta-yaw rotational velocities.

To this end, series of experiments are conducted to evaluate the frameworks performance in satisfying the pointed key optimization criterion. Accordingly, the UUV's maximum velocity is set to 5.5 knots (~2.82m/s) assuming the use of standard propellers; the battery life-time is assumed 4 hours (14,400 sec) for a continuous operation; the dimension of operation field is considered in a scale of ($10_x \times 10_y \times 1_z$) kilometres; total number of stations, including fixed and moving ones, is set on 20 nodes; the start and goal stations are defined as node-1 and node-20, respectively. Presuming the use of on-board sonar and HADCP sensors for continuous measuring of the local environmental updates, the water current field is computed from a random distribution of 2 to 5 Lamb vortices in a 200×200 grid within a $2_x \times 2_y$ kilometres spatial domain while a Gaussian noise, in a range of 0.1~0.8, is randomly applied to update current parameters of $\Gamma^o$, $\ell$, and $\Im$ given in (5). The goal is to determine how the DE algorithm can satisfy the given objectives in a hazardous dynamic environment. Therefore, the following plots are produced by implementing the system in MATLAB®2016 to present the model performance in satisfying the key criterion of reactive local-global planners. Fig.8 depicts performance of the local-planner in coping with current behaviour (using the favourable current and avoiding vortices in contrast). As can be seen current deformation is covered in each path section. Being more specific, the proposed local planner shows significant flexibility in adapting current deformations specifically when the magnitude gets sharper (close to vortices).

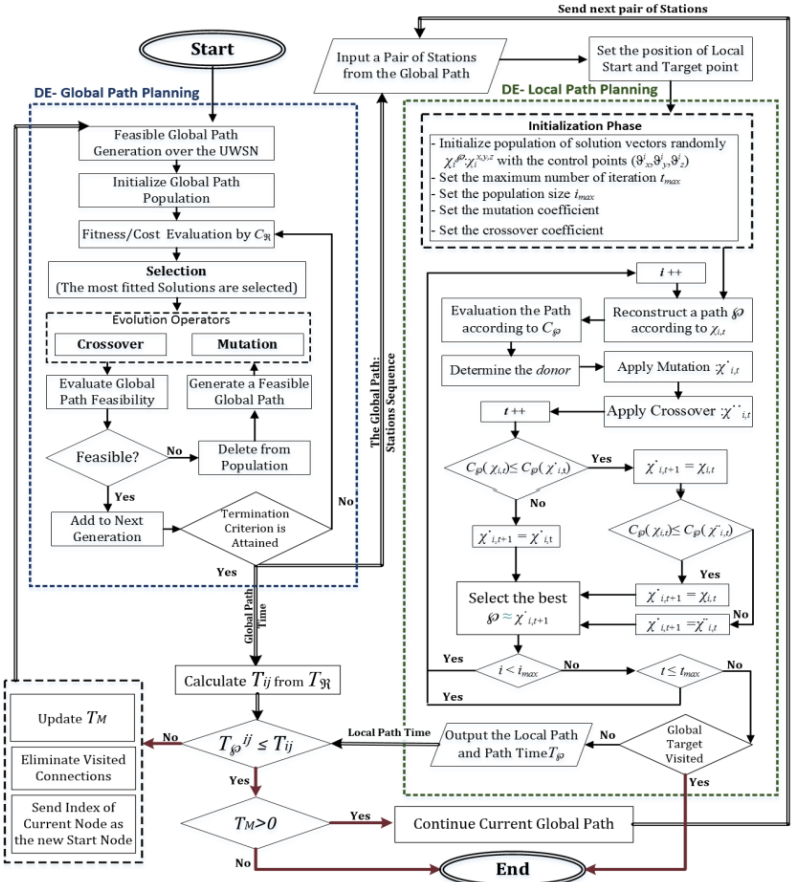

Fig.7. Process of the introduced DE based multilayered motion planning framework.

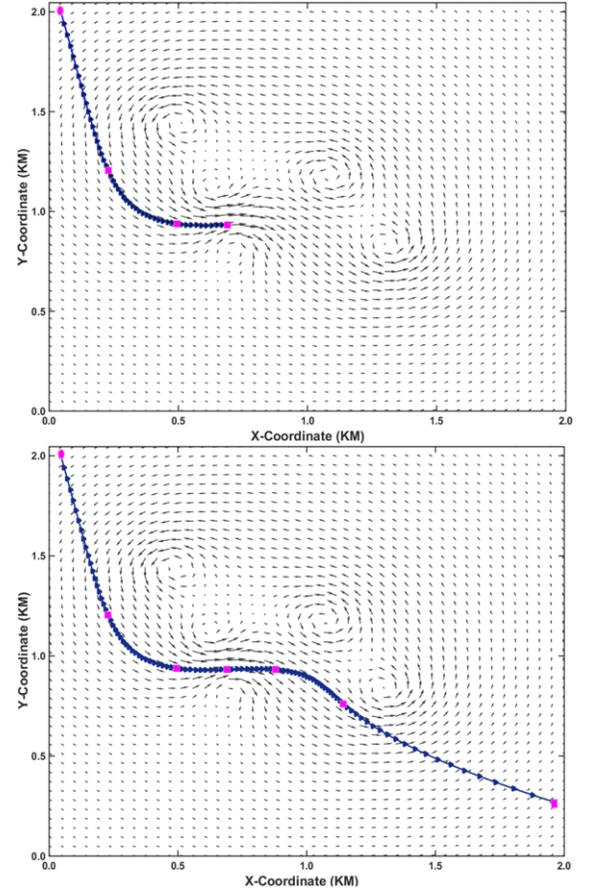

Fig.8. Adaption of the local path to the current transformation.

The performance of the framework in a UUV's single mission is proposed by Fig.10 in which rearrangement of order of stations in a global path according to updated mission time $T_M$ and networks deformations is presented through three global re-planning procedures. In this process, the local path also re-planned for several times due to copying with environmental local changes such as collision avoidance. In the beginning of the mission, the base layer produces a global path fitted to $T_M$.

Referring to Fig.9 and Table.1, the initial global path takes $T_\Re=13,983$ (*sec*) duration (which is smaller than $T_M=14,400$ (*sec*)) and encapsulates number of 11 stations (including {1-9-12-8-15-3-7-13-5-18-20}) with total value of 46. The inner layer then, tends to generate feasible and efficient local path between stations. As presented by Fig.9(a), the local planner refines the path between the first pair of stations {1-9} to avoid colliding a moving obstacle, which cause a time dissipation. After each run of the inner layer, the network deformation is considered, the local path time $T_\wp$ is compared to expected time and then $T_\wp$ is reduced from the total existent time $T_M$. If the local path $\wp$ took longer than expectation ($T_\wp^{ij} > T_{ij}$) the global re-planning flag is triggered to compensate the time dissipation, which is the case in the second pair of stations {9-12} in the initial global path. Hence, the base layer is recalled to generate the second optimal global path fitted to the updated $T_M$.

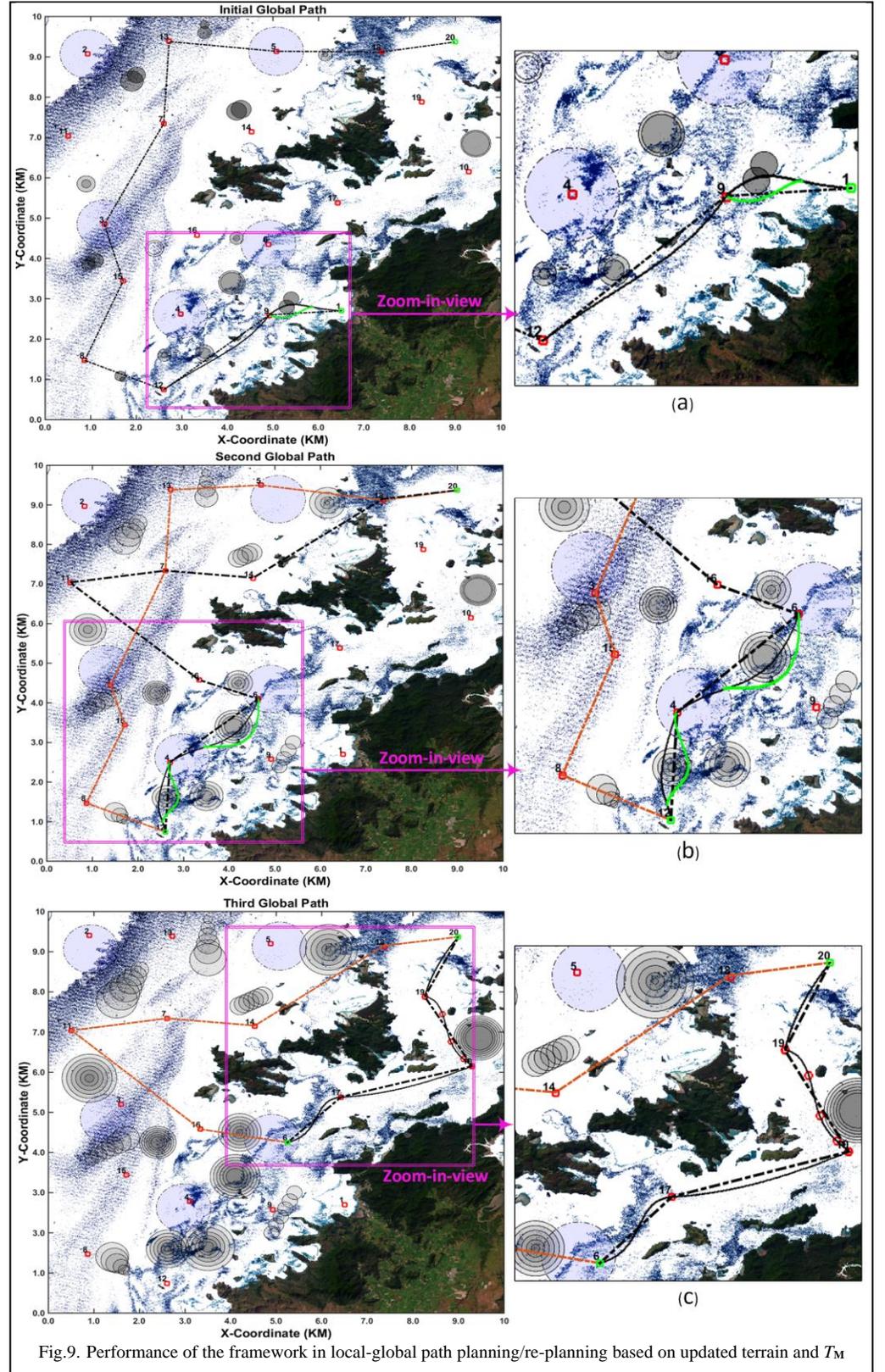

Fig.9. Performance of the framework in local-global path planning/re-planning based on updated terrain and $T_M$

Referring to Fig.9(b), the second global path starts from the existing station (node-12) and includes 9 stations {12-4-6-16-11-7-14-18-20} with the total value of 43. According to the given information in Table.1, the second global path takes $T_\Re=11,002$ (*sec*) duration, which is smaller than updated $T_M=11,347$ (*sec*). Similarly, the node sequence is shifted to the inner layer and the process of local path planning is repeated till the $T_\wp^{ij}$ exceeds the $T_{ij}$, or the existing global path time $T_\Re$ exceeds the $T_M$ due to the network deformations. The local path planner finishes its job on time when arrives to station-4, but its process gets delayed in distance between

node-6 to node-16. Hence, another global path is re-planned to cover the loos of time, which is presented by Fig.9(c). The third global path involves 5 stations {6-17-10-19-20} with the value of 28. As depicted by zoom-in-view in Fig.9(c) the local planner accurately avoids colliding obstacles. It operates continuously without any delay and the vehicle reaches on-time to the target node-20 by remaining time of 213 (*sec*). In Fig.9, the purple circles around some of the stations represents the motion board of the floating moored sensors/stations. As can be seen in the sequence of plots in Fig.9, the floating stations change their positions over the time. The black dashed line represents the current valid global path while the orange dashed line represents the discarded global path. Similarly, the black thick lines represent the local paths generated by the inner layer, and the green thick lines are the refined paths. The transparent black circles around the obstacles represent the collision boundaries with the confidence of 98% that the object is located within this area with an uncertainty propagation (presented by gradual increment of the circles). The overall mission involves 3 global and 3 local path re-planning. The final travelled path includes 9 stations of {1-9-12-4-6-17-10-19-20}, with the total value of 41. No collision occurred over the operation and mission terminated successfully with no delay (with remaining time of $T_M$=14400-14187=213 (*sec*)), which means completion of the mission before running out of battery. The given data in Table 1 declares the feasibility of the generated global path, while it is satisfying the time constraint. It is also notable that the global path time index is optimized as the vehicle has used the maximum of existent $T_M$ to meet topmost possible stations.

TABLE I
THE PERFORMANCE THE FRAMEWORK IN A SINGLE MISSION (GIVEN BY FIG.9).

| Global Path | First | Second | Third | Final Resultant Global Path |
|---|---|---|---|---|
| Time ($T_R$) (sec) | 13983 | 11,002 | 658 | 14,187 |
| Existent Time ($T_M$) (sec) | 14,400 | 11,347 | 856 | 14,400 |
| Total Value of Stations | 46 | 43 | 28 | 41 |
| Number of Stations | 11 | 9 | 5 | 9 |
| Cost ($C_R$) | 1.093 | 1.149 | 1.071 | 1.031 |
| CPU Time (sec) | 283 | 240 | 183 | 266 |

The local path planner's performance in satisfying Surge-Sway-Yaw rate constraints is also investigated and performed by Fig.10. According to (15), the kinematic constraints of the UUV are defined as follow: the surge velocity of *u* should not exceed $v_{x,max}$=2.7(*m/s*); the sway velocity of *v* should be bundled in [$v_{y,min}$=-0.5(*m/s*), $v_{y,max}$=0.5(*m/s*)]; and the rotational velocity of yaw rate should be limited to $\dot{\psi}_{min}$=-17 (*deg/s*) and $\dot{\psi}_{max}$=17 (*deg/s*) in all states along the local path. Here, the red dashed horizontal lines show the violation boundaries for all above parameters.

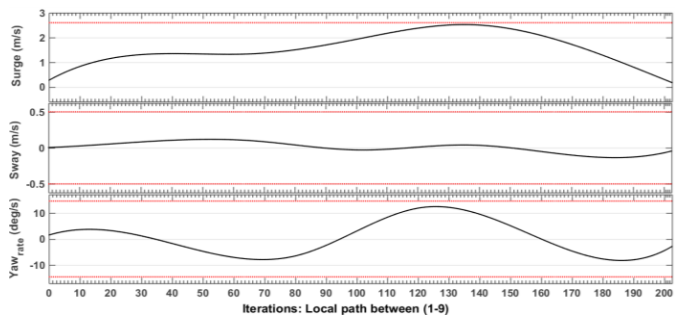

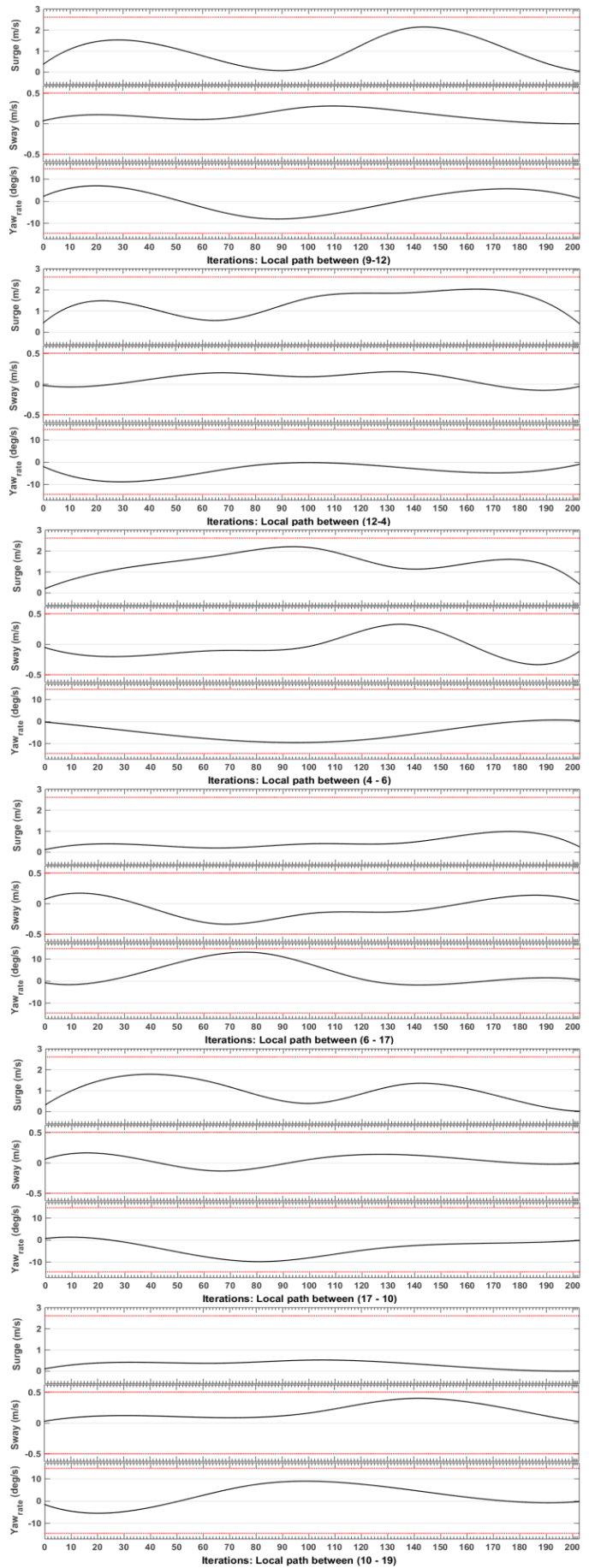

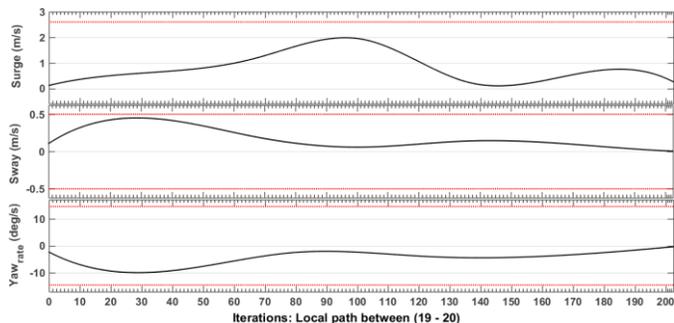

Fig.10. Variations of Surge, Sway, and Yaw parameters for the local path over the time (iterations).

The black lines in Fig.10 represent the UUV's velocity and orientation at each single spot on the produced local path curve in which the horizontal axis is in the same size of the B-Spline vector. It is evident from the simulation given by Fig.10, all local paths in the mission efficiently satisfy the defined kinematic constraints of the UUV as all parameters are lied in the valid boundaries.

Fig.11 shows the cost variation for local-global planners and the entire framework, in which cost for the local planner is function of local path time-distance and vehicular-collision violations (see (15)); cost of the global planner is a constrained function of global path time and stations value (see (10)); and the total model cost is a combined function of local and global path time (see (16)). As can be seen, the algorithm tends to minimize the cost iteratively. The behaviour of the framework in satisfying addressed objectives is investigated in a quantitative manner through the 30 experimental trials (presented by Table.2), where the topology of UWSN is randomly transforms with Gaussian distribution in the problem search space. It is outstanding from all experiments in Table.2, the residual time got a very small positive value comparing to total battery life-time (14,400 $sec$) that means the UUV completed its mission in available battery life-time and no failure is occurred in presented experimental trials.

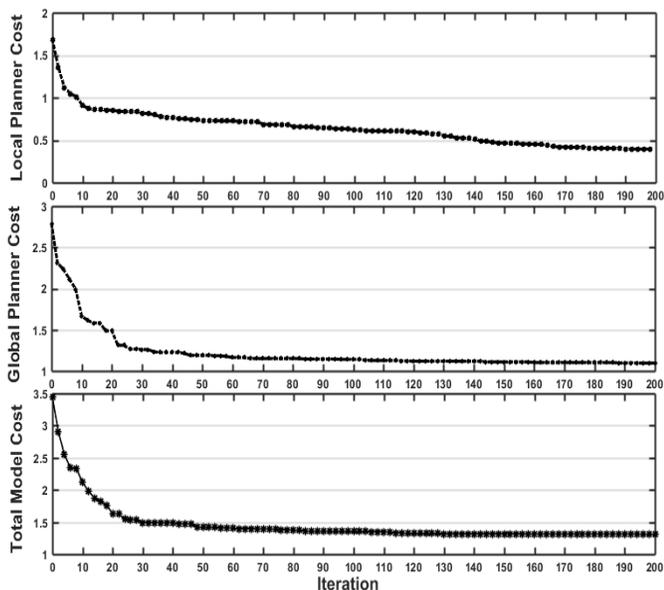

Fig.11. Cost variations of the local planner, global planner and the entire multilayered framework in a single execution.

TABLE II. THE PERFORMANCE THE DE BASED MULTILAYERED FRAMEWORK IN MULTIPLE EXPERIMENTS

| Experiment # | Global Re-plan # | Global Path Time ($sec$) | Residual Time ($sec$) | Total Value of Stations | Visited Stations # | Total Cost | CPU Time ($sec$) |
|---|---|---|---|---|---|---|---|
| 1 | 3 | 14187 | 213.0 | 41 | 9 | 1.031 | 266 |
| 2 | 5 | 13956 | 444.0 | 38 | 8 | 1.233 | 348 |
| 3 | 3 | 13868 | 532.0 | 33 | 7 | 1.396 | 214 |
| 4 | 4 | 14348 | 52.00 | 43 | 11 | 1.012 | 254 |
| 5 | 6 | 13792 | 608.0 | 42 | 11 | 1.151 | 376 |
| 6 | 3 | 14012 | 388.0 | 39 | 9 | 1.336 | 253 |
| 7 | 4 | 13971 | 429.0 | 40 | 11 | 1.213 | 288 |
| 8 | 4 | 13906 | 494.0 | 36 | 8 | 1.306 | 293 |
| 9 | 7 | 13783 | 617.0 | 30 | 6 | 1.543 | 384 |
| 10 | 3 | 14324 | 76.00 | 49 | 15 | 1.113 | 246 |
| 11 | 5 | 13898 | 502.0 | 44 | 14 | 1.243 | 303 |
| 12 | 6 | 13821 | 579.0 | 43 | 12 | 1.301 | 316 |
| 13 | 5 | 13846 | 554.0 | 37 | 8 | 1.472 | 328 |
| 14 | 4 | 14003 | 397.0 | 39 | 8 | 1.335 | 303 |
| 15 | 3 | 14362 | 38.00 | 45 | 10 | 1.019 | 291 |
| 16 | 3 | 14235 | 165.0 | 38 | 9 | 1.114 | 246 |
| 17 | 4 | 14154 | 246.0 | 36 | 7 | 1.103 | 311 |
| 18 | 3 | 14289 | 111.0 | 46 | 11 | 1.096 | 296 |
| 19 | 6 | 13874 | 526.0 | 40 | 9 | 1.218 | 371 |
| 20 | 4 | 14333 | 67.00 | 48 | 13 | 1.019 | 331 |
| 21 | 5 | 13993 | 407.0 | 34 | 8 | 1.346 | 346 |
| 22 | 2 | 14250 | 150.0 | 41 | 12 | 1.097 | 298 |
| 23 | 4 | 13843 | 557.0 | 39 | 11 | 1.183 | 302 |
| 24 | 4 | 13897 | 503.0 | 42 | 11 | 1.168 | 316 |
| 25 | 2 | 14382 | 18.00 | 51 | 15 | 1.001 | 245 |
| 26 | 3 | 14014 | 386.0 | 47 | 9 | 1.091 | 286 |
| 27 | 5 | 13769 | 631.0 | 36 | 8 | 1.536 | 354 |
| 28 | 4 | 14023 | 377.0 | 38 | 8 | 1.344 | 329 |
| 29 | 6 | 13853 | 547.0 | 35 | 9 | 1.406 | 381 |
| 30 | 3 | 14326 | 74.00 | 46 | 14 | 1.179 | 276 |
| STD | 206.39 | 206.39 | 854.14 | 204.47 | 0.1562 | 44.354 | |
| M ± SE | 14043.73 ±37.6 | 356.26 ±37.6 | 198.96 ±158.6 | 47.93 ± 37.9 | 1.22 ±0.02 | 305.03 ± 8.09 | |

Minimizing the residual time means taking maximum use of existent time, which is important factor for having a reliable and efficient mission. Consequently, the analysis of 100 Monte Carlo simulation on cost variations is provided by Fig.12. Monte Carlo simulations investigates the efficiency of the algorithm in dealing with random transformation of the UWSN topology, where the number of stations vary with a uniform distribution in {20, 50} and the UWSN topology transforms with Gaussian distribution in the problem search space.

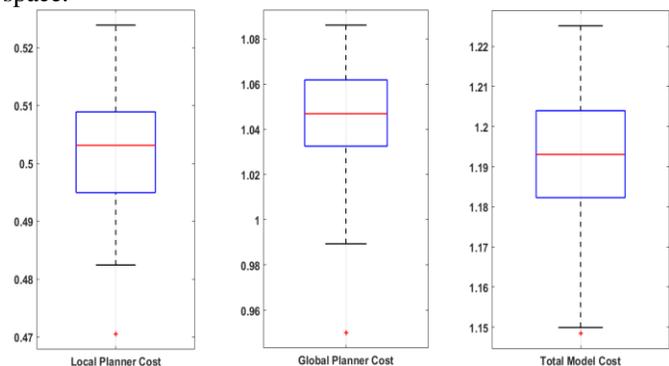

Fig.12. Cost variations of local-global planner and the entire model in 100 Monte Carlo simulations.
11

It is outstanding from Fig.12, the cost value for the local path planner is approximately drawn around (0.49,0.51), where this variation is around (1.03,1.07) for the global planner and (1.17,1.22) for the entire multilayered framework. As can be seen in Fig.12, the generated solutions are consistently distributed considering problem space deformation. The obtained statistical results demonstrate the inherent robustness and stability of the proposed strategy in a quantitative manner.

Eventually, it is perceived from demonstrated simulation results in Figs.8-12, the introduced multilayer framework accurately manages the operation time by producing efficient global path in the base layer while dealing with UWSN deformations; and concurrently deals with environmental immediate changes (such as current deformations and obstacle appearance) using the inner-layer local path planner. The applied algorithm is accurate against current immediate updates and its re-planning functionality facilitates the vehicle to simultaneously refine its trajectory to avoid collision and turbulent areas. The re-planning procedure uses the useful information of previous solutions to re-produce a new trajectory and this fact decreases computational burden of path planning. The results of simulations indicate that the DE algorithm is able to meet the constraints and to optimize the performance index, which defined by equations (15) and (16). More precisely, Fig.11 shows the fast convergence rate of feasible solutions obtained over 200 iterations by the DE. With respect to above discussion, DE seems a logical and appropriate choice for using in the designed multilayered motion planner.

## V. Conclusion

A multilayered motion planner with a mounted energy efficient local path planner is introduced in this study for a UUV navigation throughout a complex and time variant semi-dynamic UNSW. The UUV routing protocol in the network is generalized with a Dynamic Knapsack-Traveler Salesman Problem (DKTSP) along with an adaptive path planning problem to address UUV's long-duration missions on dynamically changing subsea volume. Mobile stations' (buoyant wireless sensors) position and status change randomly over the time and space. These sensors forward collected information to the vehicle along its travel toward the target node. The global path planner is integrated with a current resilient local path planner and accommodates the UUV to serve maximum possible mobile-fixed stations of the network while the inner layer generates energy efficient trajectories by using desirable and avoiding undesirable current flow. Making use of desirable water current propels the vehicle along its trajectory and leads saving more energy. This considerably reduces the total UUV mission costs and facilitate the vehicle to handle longer operations. The global path planner in the base layer splits the operating field to the smaller operation windows bounded to the pairs of sensors, which eases data collection and data analysis for the inner layer. This remarkably accelerates the re-planning process as less data is required to be rendered and recomputed. The DE algorithm is used to evaluate the performance of the multilayered framework in satisfying mission objectives. Owing to real-time capability of the DE algorithm and reactive multilayered structure of the proposed framework, the simulation trials have demonstrated efficient mission management performance even considering some constraints on vehicles kinematic and environment, etc. The framework guarantees on-time mission completion by an appropriate ordering of the stations in the global path; and assures a safe and efficient deployment by avoiding severely adverse current flows and using favorable ones to propel the vehicle and reduce energy expenditure.

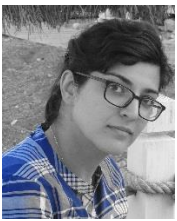

**Somaiyeh MahmoudZadeh** completed her PhD at School of Computer Science, Engineering and Mathematics, Flinders University of South Australia. Her area of research includes computational intelligence, autonomy and decision making, situational awareness, and motion planning of autonomous underwater vehicles.

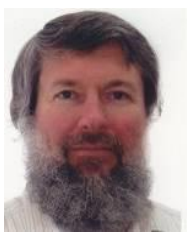

**Prof. David M W Powers** is Professor of Computer Science and Director of the Centre for Knowledge and Interaction Technology and has research interests in Artificial Intelligence and Cognitive Science. He is known as a pioneer in Parallel Logic Programming, Unsupervised Learning and Evaluation of Learning, and was Founding President of ACL SIGNLL.

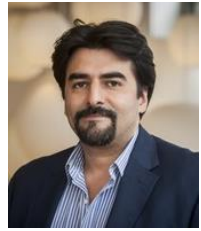

**Dr. Adham Atyabi** received his PhD from Flinders University of South Australia. He is currently acting as Technology Lead in Seattle Children's Innovation & Technology Lab and Senior Postdoctoral Fellow at University of Washington. His research interests include Brain Computer Interfacing, Machine Learning, Cognitive Robotics, and Evolutionary Optimization.